\documentclass[sigconf]{aamas} 

\usepackage{balance} 
\usepackage{booktabs} 
\usepackage{paralist}
\usepackage{xspace}
\usepackage{gensymb}
\usepackage{graphicx}
\usepackage{tabularx}

\setcopyright{ifaamas}
\acmConference[AAMAS '23]{Proc.\@ of the 22nd International Conference
on Autonomous Agents and Multiagent Systems (AAMAS 2023)}{May 29 -- June 2, 2023}
{London, United Kingdom}{A.~Ricci, W.~Yeoh, N.~Agmon, B.~An (eds.)}
\copyrightyear{2023}
\acmYear{2023}
\acmDOI{}
\acmPrice{}
\acmISBN{}

\acmSubmissionID{59}

\title[Which Way Is 'Right'?]{Which way is `right'?: Uncovering limitations of Vision-and-Language Navigation models}

\author{Meera Hahn}
\affiliation{
  \institution{Google\\Georgia Institute of Technology}
  \country{United States}}
\email{meerahahn@google.com}

\author{Amit Raj}
\affiliation{
  \institution{Google\\Georgia Institute of Technology}
  \country{United States}}
\email{amitrajs@google.com}

\author{James M. Rehg}
\affiliation{
  \institution{Georgia Institute of Technology}
  \country{United States}}
\email{rehg@gatech.edu}

\begin{abstract}
The challenging task of Vision-and-Language Navigation (VLN) requires embodied agents to follow natural language instructions to reach a goal location or object (e.g. `walk down the hallway and turn left at the piano'). For agents to complete this task successfully, they must be able to ground objects referenced into the instruction (e.g.`piano') into the visual scene as well as ground directional phrases (e.g.`turn left') into actions. In this work we ask the following question -- to what degree are spatial and directional language cues informing the navigation model's decisions? We propose a series of simple masking experiments to inspect the model's reliance on different parts of the instruction. Surprisingly we uncover that certain top performing models rely only on the noun tokens of the instructions. We propose two training methods to alleviate this concerning limitation.
\end{abstract}


\newcommand{\BibTeX}{\rm B\kern-.05em{\sc i\kern-.025em b}\kern-.08em\TeX}
\newcommand{\myquote}[1]{\emph{`#1'}}
\newcommand{\xhdr}[1]{\vspace{3pt}\noindent\textbf{#1}}

\usepackage[utf8]{inputenc}
\begin{document}
\pagestyle{fancy}
\fancyhead{}
\maketitle 
%

\section{Introduction}
Vision Language Navigation (VLN) is the task of having a robot navigate a visual 3D environment via following human generated natural language instructions such as \myquote{Leave the bathroom and walk to the right}. VLN is a popular task with multiple benchmark datasets~\cite{anderson2018vision,ku2020room,chen2019touchdown,qi2020reverie,thomason2019vision} which are largely conducted in simulated indoor environments such as Matterport3D (MP3D) ~\cite{chang2017matterport3d} and contain thousands of human annotated instructions. MP3D is constructed of a set panoramic nodes connected via navigational ability to form a navigation graph. The task is challenging as it requires accurate visual grounding of objects and visual descriptions provided in the instructions into the environment. Furthermore it requires the model to understand spatial language to ground instructions such as \myquote{walk to the right} into actions. The action space at a given time step for the navigating agent is to move to a neighboring node or to cease navigation. 

Solutions to the VLN task can be divided into two distinct settings: a discriminative path-ranking setting and a generative path selection setting. In the path-ranking setting, using beam search from the given starting location, up to 30~\cite{majumdar2020improving, guhur2021airbert} possible paths are generated and a path is selected using a discriminative path selection model. In the generative setting the agent is placed at the starting location and the model sequentially selects the next node to navigate to, until the model selects the stop action. 

Both sequential and path-ranking settings have seen large success in modeling by using multi-modal transformer models which leverage large-scale pre-training~\cite{majumdar2020improving, guhur2021airbert, hong2021vln}. The data used to pre-train these models consists of large scale web data~\cite{sharma2018conceptual,murahari2020large,lu2019vilbert,hao2020towards} containing image-text pairs to learn visual grounding as well as large text corpora~\cite{zhu2015aligning} to learn linguistic semantics. 

In many instances in the R2R dataset, the instructions dataset refer to the spatial layout of objects in the environment and the agents position to these objects. For example ``walk past the green sofa in front of you and turn right.'' Due to the nature of the nodes being panoramic, learning the meaning of ``in front'' as well as then relating this to the ``right'' direction is challenging. In this paper we seek to understand to what degree the model is learning spatial and directional words and how these words impact the performance of the model. Prior work~\cite{wang2021less, zhao2021evaluation, zhu2021diagnosing, zhang2020diagnosing,thomason2018shifting} has investigated the failure modes of sequential models. These works find that when making predictions, sequential agents attend equally to object and direction tokens in the navigation instruction. However, there is minimal diagnostic evaluation over the path-ranking VLN models.

In this paper we outline a simple method via token masking to understand how different types of part of speech and object vs direction tokens are used by path-ranking models. Through our experiments we find that the path-ranking models rely most heavily on nouns, almost disregarding direction tokens and other parts of speech. Additionally, we find that changing direction tokens while holding nouns tokens constant leads to no effect on the model predictions. This highlights a large limitations of these models as they are not capturing large amounts of the available information for predictions. We investigate two different training procedures in an effort to alleviate the models reliance on nouns. We hope that these findings which reveal the limitations of current VLN models will lead to new research.\\

\noindent\textbf{Contributions}:
\begin{compactenum}
    \item We develop a diagnostic procedure to determine the influence of tokens types on VLN model's predictions. 
    \item We uncover the phenomena that path-ranking VLN models are almost only attending to noun tokens and completely disregarding spatial tokens and all other parts of speech. 
    \item We propose 2 training procedures for VLN-Bert to alleviate the affects of noun token reliance as well as increase overall accuracy. 
\end{compactenum}
\section{Background}
\label{sec:background}

\subsection{Embodied Language Tasks.} A number of Embodied Artificial Intelligence (EAI) tasks which involve natural language, visual perception, and navigation in realistic 3D environments have recently gained prominence. These tasks include Embodied Question Answering~\cite{das2018embodied,gordon2018iqa}, instruction based navigation~\cite{anderson2018vision, chen2019touchdown, mehta2020retouchdown, qi2019reverie, chaplot2018gated}, and household cleaning and organization tasks~\cite{puig2018virtualhome,ALFRED20}. EAI involves active perception -- performing physical actions determined via linguistic or visual observations. Within the larger umbrella of instruction-following there are multiple tasks which differ in: the way the goal is specified, types of environments, number of agents, and availability of Oracle assistance. Specifically, Vision-Language Navigation (VLN)~\cite{anderson2018evaluation, ku2020room} is a foundational task in which an agent is given natural language instruction to follow to a goal location in a novel environment. Key challenges of the task include: accurate language grounding, efficient navigation and generalization to novel environments, objects and language. Note the unseen environments and instructions contained in the test split, can contain language and objects that were not present in the training episodes. 

Following works~\cite{nguyen2019help, thomason2019vision} extend the VLN task to allow an agent to query an Oracle for assistance in the form of further fine-grained instructions and dialog~\cite{thomason2019vision, hahn2020way}. Ku et al.~\cite{ku2020room} extend the original VLN benchmark dataset with additional languages, more variable paths and pixel-wise groundings for instructions. ~\cite{qi2019reverie, chen2019touchdown} extends the VLN paradigm by adding referring expressions to specific objects that the agent must locate at the goal location. In this work we specifically look at the version of VLN task that does not allow pre-exploration of the environment~\cite{wang2020active}, meaning that, each environment is completely novel to the agent when it starts navigating.

\subsection{Pre-training VLN Models.}
Current VLN datasets are small in size, as seen in Table \ref{table:dataset_comparison}, due to size limitations of existing simulation datasets and the high cost of human annotations. To combat this challenge VLN methods have relied on data augmentation strategies, auxiliary tasks and extensive pre-training schemas. A common form of data augmentation in VLN are instructions generated by a trained speaker model~\cite{fried2018speaker, ma2019regretful, tan2019learning}. Examples of training with auxiliary losses includes ~\cite{hao2020towards, huang2019multi} which trained with path-instruction compatibility as an auxiliary loss. The majority of recent methods have sought to distill language grounding into the model via pre-training on disembodied web data~\cite{majumdar2020improving, hahn2022transformer, guhur2021airbert}. In this work we specifically examine three alternate pre-training procedures in an effort to alleviate the model's limitations that were exposed in the masking experiments.

\subsection{Analysis of Visiolinguistic Models.}
As an effort to improve and understand the capabilities of instruction following agents, prior work has investigated the limitations of both VLN models and the training procedures of the models. Two works~\cite{zhao2021evaluation, zhang2020diagnosing} have investigated the limitations of the commonly used speaker-follower model. A speaker model is used as a data augmentation method by generating more instructions over the paths in the VLN datasets. Zhao et al.~\cite{zhao2021evaluation} perform an extensive study in which human wayfinders are used to evaluate the instructions generated by the speaker model and score compatibility with the true path. They uncover that most speaker model generators perform equivalent to a template-based generator and extremely worse than human instructor. Additionally they discover that with the exception of SPICE, common text similarity metrics (BLEU, ROUGE, METEOR and CIDEr) are ineffective for evaluating generated instructions against a reference instruction. ~\cite{wang2021less} also find that speaker models suffer from poor visual grounding and create inadequate instructions. To mediate this, they propose a two stage generation model using visual landmark detection and an encoder-decoder framework.

Multiple works~\cite{zhang2020diagnosing, liu2021vision} have presented analysis of the key challenge in the VLN task: generalization to unseen environments. As previously discussed, VLN datasets are small in terms of number of environments and number of instructions. This has lead to large bias towards training environments which ~\cite{zhang2020diagnosing} studies via a set of diagnostic experiments on SOTA methods to find the specific underlying causes of the bias. They find that the low-level visual information, contained in the ResNet features, contributes substantially to environment bias. They propose several alternate feature representations which lead to a drop in domain shift between train and test environments. Closest to the work presented in our paper, ~\cite{zhu2021diagnosing} sought to understand how instruction following agents perceive the multi-modal inputs. Similar to our work, they designed experiments to unveil the focus of the agent over the input while making navigation decisions. However, ~\cite{zhu2021diagnosing} only studied sequential based instruction following models. They found that, out of the models they examined, indoor navigation agents refer to both object and direction tokens when making predictions; however, outdoor navigation agents heavily use direction tokens and poorly understand the object tokens. Via ablations on visual tokens and text tokens, \cite{zhu2021diagnosing} find ablating visual tokens has a significantly smaller impact on navigation performance than the text object ablations therefore indicating an unbalanced attention of the vision and text inputs. In contrast to Zhu et al.~\cite{zhu2021diagnosing}, our paper focuses on exposing the limitations of path-ranking VLN models. We don’t study outdoor navigation in this paper because, to our knowledge, path-ranking models are not used as an approach for outdoor navigation however they heavily dominate the indoor instruction-following space. Thomason et al.~\cite{thomason2018shifting} additionally investigates the multi-modality of SOTA models by ablating SOTA models per modality input. The work concludes that SOTA models with access to only a single modality achieve surprisingly high results, beating out baselines with access to both modalities. The authors suggest using uni-modal ablations as a best practice for future methods to understand bounds of performance related to uni-modal biases in multi-modal datasets.

\subsection{Analysis on Instructional Language of Benchmark Datasets}
\label{comp_language}
In Table \ref{table:dataset_comparison} we compare the language of instructions between common VLN benchmarks~\cite{qi2020reverie, ku2020room, anderson2018vision} as a means to understand the composition of token types in the navigational instructions. We compare the sizes of the datasets and measure the density of different token sets per episode. The vocab size of a dataset is the total number of unique tokens. The average text length per episode is simply the average length of instructions in the dataset. In measuring density across different token sets of the dataset, we specifically look common parts of speech (POS), object, numeric and spatial tokens.  We calculate density this by running a POS tagger~\cite{loper2002nltk} over all the instructions. Object and spatial token sets are pre-defined and described in Section ~\ref{methods}. Note that in Table \ref{table:dataset_comparison} Left-Right refers to the combined density of the words `left' and `right'. 

Looking at Table \ref{table:dataset_comparison}, we first observe the overwhelming presence of nouns in the instructions. 
Nouns have the highest density of all POS. For instance, in the Reverie dataset, the density of nouns is 5 times greater than that of adjectives. This is unsurprising as it follows larger trends in the English language \cite{liang2013noun}. We observe a low density of spatial words like ``left'' and ``right''. This is expected as these words are assumed to be only used a few times per instruction. We note that the density of object tokens is very similar to the density of spatial tokens and is smaller than of the density of noun tokens by 50\%. 

\begin{table}[htp]
\begin{center}
\caption{Comparison of the language in common Vision-Language Navigation benchmark datasets. All datasets contain triplets of starting location, goal location and a natural language instruction. Here we compare the size of the datasets and density of different parts of speech in the instructions. Density refers to the average percentage of tokens in a instruction which are of that part of speech. \textit{Density: Left-Right} is the combined density of the words \textit{left} and \textit{right}.}
\label{table:dataset_comparison}
\resizebox{\columnwidth}{!}{
\begin{tabular}{lccc} 
\toprule
 &  Reverie~\cite{qi2020reverie} &  RXR~\cite{ku2020room} &  R2R~\cite{anderson2018vision} 
\\ \toprule
\# Instructions & 22k & 126k & 22k \\
\#  Paths & 7K & 16.5k & 7k \\
Vocab Size $^{\dagger}$ & 4815 & 3779 & 3999 \\
Avg Instr Length $^{\dagger}$ & 18.31 & 97.30 & 29.37 \\ \midrule
Density: Noun $^{\dagger}$ & 0.32 & 0.21 & 0.28 \\
Density: Adjective $^{\dagger}$ & 0.05 & 0.06 & 0.05 \\
Density: Verb $^{\dagger}$ & 0.12 & 0.17 & 0.12 \\
Density: Objects$^{\dagger}$ & 0.15 & 0.07 & 0.10 \\
Density: Numerical$^{\dagger}$& 0.02 & 0.01 & 0.01 \\
Density: Spatial$^{\dagger}$ & 0.10 & 0.10 & 0.13 \\
Density: Left-Right$^{\dagger}$ & 0.05 & 0.05 & 0.07 \\ \bottomrule 
\end{tabular}}
\footnotesize{\\$\dagger$ Represents statistics which were only measured on English instructions. }\\
\end{center}
\end{table}

\section{Masking Experiments}
\label{methods}

\subsection{Methods}
\xhdr{Research Questions.}\\
\noindent In this section we aim to answer the following research questions:
\begin{enumerate}
    \item Does there exist a specific set of linguistic cues in the instructions which more heavily inform navigational decisions?
    \item Alternatively, are there sets of linguistic cues in the instructions which the agent largely ignores when making navigational decisions?
    \item Directional cues differentiate navigational instructions from other tasks. How much are instruction following agents learning and attending to the spatial and directional cues present in the instructions?\\
\end{enumerate} 

\xhdr{Ablation Experiment Design.}\\
\noindent To answer these questions, we create a set of ablation experiments over the navigation instructions and evaluate on standard trained SOTA VLN models. Specifically we modify the navigational instructions by removing (via masking) or replacing tokens of a specific linguistic cue set which fall into a particular part of speech (POS) or if they are a object/spatial/numeric token. Additionally we test the uni-modal ablation of masking an entire modality of language or visual input. We examine each ablation experiment effects on the agents’ evaluation performance while keeping all other variables unchanged. We chose to partition text tokens into different cue sets by their part of speech as well as if they fall into a pre-defined set of directional words.

\begin{figure*}[!ht]
\centering
\includegraphics[width=\textwidth]{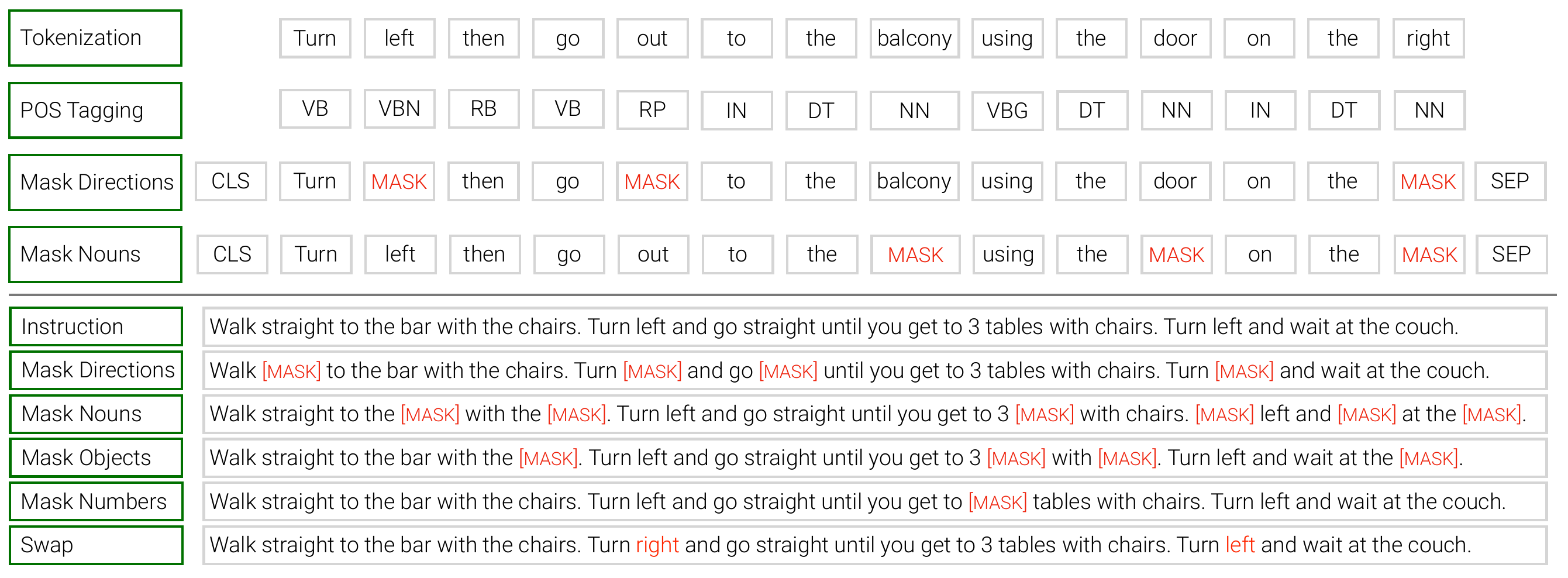}
\caption{Example augmentation of navigational instructions from the R2R dataset during the noun masking experiment and direction masking experiment. The instructions are first tokenized then part of speech tagged. Input tokens are masked out according to their cue set depending on the experiment criterion.}
\label{masking-illustration}
\end{figure*}

By testing the models performance while it doesn't have access to a specific type of token, we gain insight into the degree to which that type of token informs the model's predictions. We examine 5 different masking criterion: nouns, verbs, adjectives, left-right, spatial, object, numerical. There is no standard list of spatial/directional words. Therefore via qualitative analysis over the instructions contained in the standard VLN benchmarks we select the following set of tokens as spatial cues: [\textit{right, left, straight, toward, around, near, front, above, through, down, up, between, past}]. Note in the left-right masking experiment the tokens in the set [\textit{left, right}] are masked out. We add an additional experiment called \textit{swap} in which tokens in the set [\textit{left, right}] are replaced by their antonym, therefore creating a counterfactual instruction. This experiment is illustrated in Figure \ref{masking-illustration} and shows examples of the R2R datasets' VLN instructions, as well as the instruction tokenized, part of speech tagged and masked for multiple cue sets.

The intuition behind these experiments is that if the model equally attends to all types of tokens performance should drop equally between different tokens being masked. Alternatively, if a model focuses on certain cues more than others we should see an larger drop in performance when those cues are masked. Additionally, the \textit{swap} experiment quantifies the degree to which agents are following instructional phrases. I.e. if agents focus on instructional cues the replacement of the phrase `take a left' with the phrase `take a right', should have a significant impact on performance of the navigational agent. We focus our experiments on the R2R dataset for the VLN task as it is the foundational benchmark for the task and is unilingual. \\

\xhdr{Selected Models.}\\
\noindent We focus our investigation on three high performing VLN models. We take the models, trained in their standard practice, and evaluate them using the ablation experiments described above. 

The models we test are as follows: VLN-BERT~\cite{majumdar2020improving}, AirBert~\cite{guhur2021airbert}, Recurrent-VLN-BERT~\cite{hong2021vln}. VLN-BERT and AirBert are models trained and tested in the path-ranking procedure. Recurrent-VLN-BERT is trained and tested in the sequential procedure. This choice of doing the investigation over both sequential and path-ranking methodologies is integral to the comparison of the procedures. Additionally all of these models employ multi-modal transformer architectures and utilize large-scale pre-training and data augmentation techniques. Recurrent-VLN-BERT specifically utilizes augmented data from the PREVALENT~\cite{hao2020towards} speaker model during training. Additionally, AirBert is an extension of the VLN-BERT model. It leverages an additional loss and more pre-training data scrapped from Airbnb to increase performance. By comparing AirBert and VLN-BERT models in this investigation we seek to uncover information about how the Airbnb pre-training data is increasing performance. 

Table~\ref{table:human_model_comparison} shows the performance of the SOTA VLN models and compares it to the baselines of human performance and a random agent. The random agent randomly selects a direction and moves five steps in that direction. Note the significant gap that remains between the top baselines and human performance. The gap is the crux of the motivation for seeking to understand the limitations of current top performing VLN models. 

\xhdr{Evaluation Metrics.}\\
\noindent Performance for both the path-ranking and sequential VLN set up is primarily measured in terms of Success Rate (SR) which measures the percentage of selected paths that stop within 3m of the goal. Additionally, models are evaluated using success rate weighted by path length (SPL), which provides a measure of SR normalized by the ratio between the length of the shortest path and the selected path. 

Note that the in VLN the validation split is composed of two sets, val-seen and val-unseen. Val-seen refers to path-instruction pairs situated in the training environments while val-unseen (and the test split) contains episodes which are conducted in novel environments which are not contained in the training set. This composition allows the model to be tested for generalization to new environments and language. Unseen environments contain both novel and previously observed objects. Additionally the each episode in the dataset across every split is a unique path.

\setlength{\tabcolsep}{5pt}
\begin{table}[!ht]
\begin{center}
\caption{Performance of human performance and SOTA VLN models on the test split of the R2R dataset. Note the significant performance gaps between trained navigation agents and humans navigators.}
\footnotesize
\resizebox{\columnwidth}{!}{
\label{table:human_model_comparison}
\begin{tabular}{l ccc}
\toprule
Method  & NE $\downarrow$& SR $\uparrow$  & SPL $\uparrow$  \\ \toprule
PREVALENT~\cite{hao2020towards} & 5.30 & 54 & 0.51 \\
VLN-BERT~\cite{majumdar2020improving}  & 3.09 & 73 & 0.01 \\
AirBert~\cite{guhur2021airbert} & 2.58 & 78 & 0.01 \\
Recurrent-VLN-BERT~\cite{hong2021vln} & 4.09 & 63 & 0.57 \\
\midrule
Human Performance & 1.61 & 86 & 0.76 \\
Random &  9.79 & 13 & 0.12 \\
\bottomrule
\end{tabular}}
\end{center}
\end{table}

\subsection{Results and Analysis}

\begin{table}[t]
\begin{center}
\caption{Results of ablation-masking experiments: Success Rate (SR) on the val-unseen split of the R2R dataset. Evaluated on standard trained SOTA VLN models. The first row is the models performance with no augmentations over instruction input.}
\label{table:masking_results}
\begin{tabular}{l cccc}
\toprule   
Ablation && VLN-B~\cite{majumdar2020improving}& AirBert~\cite{guhur2021airbert} & Rec-VLN-B~\cite{hong2021vln}
\\ \toprule    
Original Input && 55.90 & 66.45 & 62.75 \\ 
\midrule
Nouns       && 43.68 & 49.94 & 43.93 \\
Adjectives  && 55.77 & 65.82 & 59.17 \\
Verbs       && 55.77 & 65.94 & 58.02 \\
Objects     && 50.40 & 57.47 & 54.63 \\
Numerical   && 55.94 & 63.64 & 62.91 \\
Spatial     && 55.64 & 64.58 & 49.64 \\
Left-Right  && 55.56 & 65.82 & 54.66 \\
Swap        && 56.07 & 66.03 & 46.70 \\
\bottomrule
\end{tabular}
\end{center}
\end{table}

\begin{figure}[!ht]
\centering
\includegraphics[width=\columnwidth]{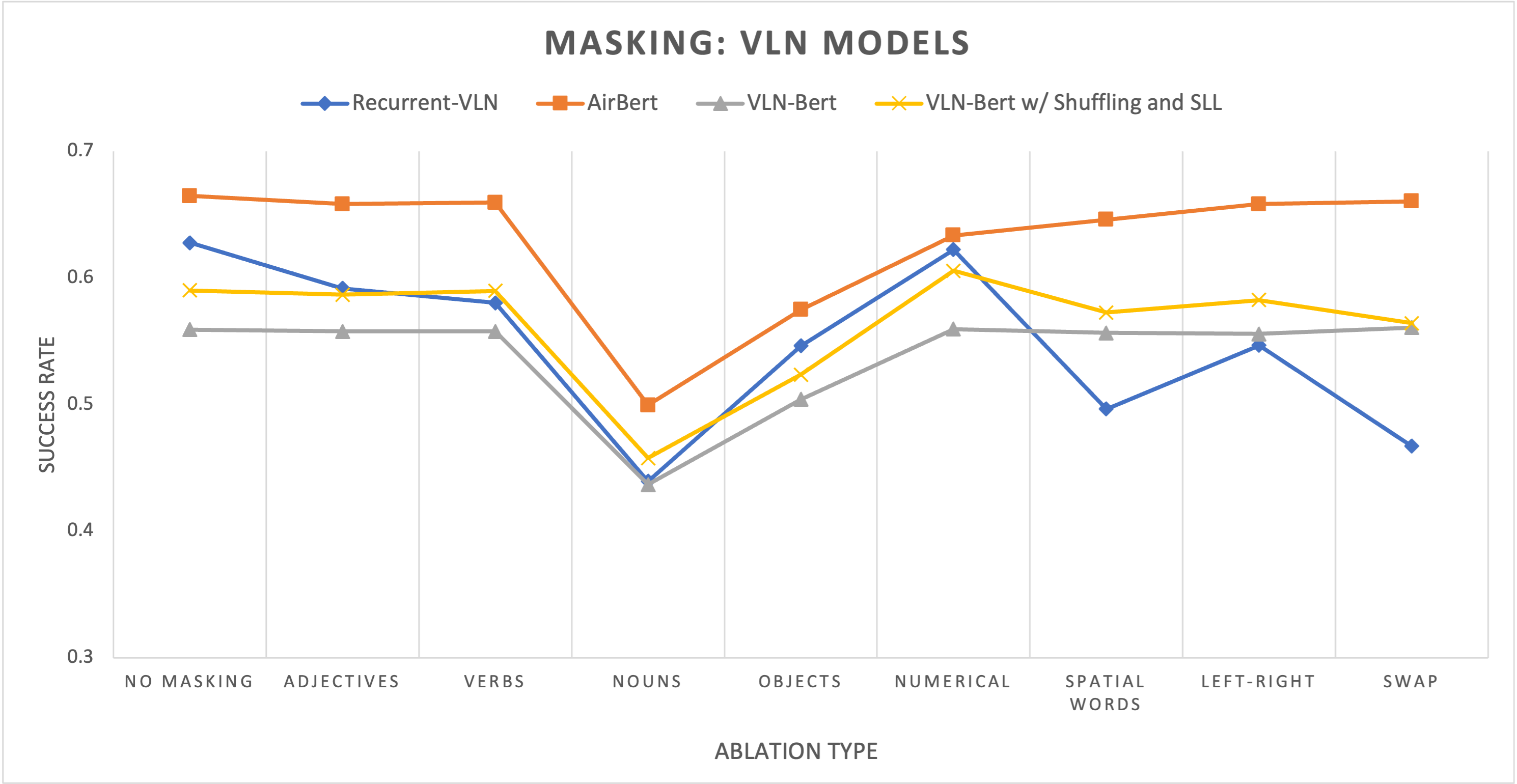}
\caption{This figure displays a visual illustration of the results of the masking experiments. Each line represents a different SOTA model along with a model (VLN-BERT w/Shuffling and SLL) that we propose in this work. We can visually see here how the ablation for nouns significantly drops performance for all models. We can also clearly identify the minimal performance change for all ablations aside from nouns for the AirBert and VLN-BERT models.}
\label{masking-vln-graph}
\end{figure}

We perform the ablation experiments on the chosen VLN models over the val-unseen split of the R2R dataset; results are shown in Table \ref{table:masking_results} and visualized in Figure \ref{masking-vln-graph}. We restate that VLN-BERT and Airbert are path-ranking models -- predicts alignment between a navigational path and a text instruction. Recurrent-VLN-BERT is a sequential model -- predicts the next node to navigate to in a iterative fashion until predicting the stop action.\\

\xhdr{Effects on Path-Ranking Models}. \\
\noindent Surprisingly, we observe in Table \ref{table:masking_results} that the performance of path-ranking models only really suffers in the case that noun tokens are masked. Performance drops less than 2\% for all other types of token masking. In fact we even observe an increase in performance for VLN-BERT by up to .0034\% when the `right' and `left' tokens are swapped to their antonym, see the \textit{swap} experiment in Table \ref{table:masking_results}. These results indicate the models heavily focus on noun tokens while making navigation decisions and seem to ignore directional information. The disregard for spatial words is concerning as they are an integral components of the navigational instructions and replacements (swaps) in directional words should result in very different paths being taken by the agent, see Section ~\ref{counterfacts} for more details. 

We posit that the path-ranking training and evaluation procedure is partially responsible for this phenomena. The inference procedure in path-ranking VLN models is that they have access to the entire navigation path when predicting alignment with the navigation instruction. We hypothesize that this framework allows the model to focus only on noun words to do pattern matching and disregard extraneous information such as positional panoramic information of the path. R2R, the dataset used in these experiments, is built over the indoor navigation environments of Matterport3D. The environments of Matterport3D range from homes to office buildings and are each very visually unique with often unique objects. In the case where there are extremely conspicuous and unique objects that serve as visual and linguistic landmarks along a path, the need to understand and regard spatial tokens becomes less necessary. We believe that within the R2R dataset and as well as other instruction following datasets set built on top of the 90 environment Matterport3D dataset -- the path-ranking VLN procedure is reducing the complexity of the task to such a degree that models can disregard most tokens in the instruction while still achieving a high success rate.  \\

\xhdr{Effects on Sequential Models.}\\
\noindent In contrast to the path-ranking methodology, we find that VLN models trained and tested with the sequential procedures take into consideration multiple types of token sets. In Table~\ref{table:masking_results} we observe that the success rate of the Recurrent-VLN-BERT model decreases under all masking conditions. We do find that masking the noun token set still creates the largest decrease in success rate. This is unsurprising as the part of speech with the highest density in the instructions is nouns, see Table ~\ref{table:dataset_comparison}. The results of the masking ablations on Recurrent-VLN-BERT align with how intuitively VLN models are expected to perform.

We posit the difference between the path-ranking and sequential inference procedures is the key factor in the difference in results between the models. The inference procedure in sequential VLN models is that the agents only have access to the neighboring environment information when predicting alignment with the navigation instruction, rather than the entire path. This makes Recurrent-VLN-BERT susceptible to cascading errors which will compound any drops in performance. Additionally as they are unable to see ahead in the path, they cannot use information about objects the instruction references that they have not seen yet to make navigational decisions. Note that this difference in task set up inherently puts sequential models at a disadvantage (in terms of success rate) compared to path-ranking models. For this reason the two types of models have rarely been compared in terms of accuracy or ablations. This novel comparison exposes shortcuts taken by path-ranking models which might suggest this procedure may need to be treated as a separate task, with a dataset tailored to be more challenging for this task. 

\xhdr{Unimodal Baselines}.
In addition to the existing masking experiments, we include uni-modal baseline ablation experiments. During evaluation, all input tokens from a single input stream are masked (either language or visual tokens). This shows the models performance while only relying on a single input modality. As the VLN task is inherently multi-modal - in that the task becomes logically infeasible given only a single modality. Therefore the performance of the model using a single modality exposes dataset bias within each modality and allows the establishment of lower bound on model performance. In Table \ref{table:masking_newmodels} we see that all VLN-BERT models retain a surprisingly high success rate when tested on uni-modal input. This indicates the current VLN datasets contain a significant bias. 

\section{Analysis of Pre-Training Procedures}
In this section we explore the question: \textit{Can we change the pre-training procedures of path-ranking models to increase their understanding and consideration of non-noun tokens?} 

To explore this questions we focus in on the VLN-BERT model which proposes a multi-stage training procedure. As the AirBert model uses a similar training procedure and architecture we believe results found on VLN-BERT can be extrapolated to the former. In this section we present three changes to the training procedures in the VLN-BERT method, with the goal of increasing the models reliance on spatial words while also maintaining performance on the R2R task. We find combining two of the methods creates noticeable changes in dependence on spatial words, while one method surprisingly produces negligible changes. 

\subsection{Data Augmentation}
Due to the small number of training and testing environments, VLN models trend to overfit to training environments and can have poor generalization to novel environments. To combat this environment bias, many VLN training methods involve extensive and creative methods of data augmentation. AirBert~\cite{guhur2021airbert} introduced a specific auxiliary training objective which intends to effectively teach the model to reason about temporal order. In this objective, the model is given an aligned path and instruction. The image tokens of the path are randomly chosen to be shuffled, and the model must predict if the path is in the correct order. We believe this training objective specifically will enforce the model to take into account the positional order of the objects in the path and possibly enforce some spatial information reliance. 

We implement a shuffling based data augmentation, inspired by AirBert, in which we create \emph{hard negative} example paths via shuffling the order of the image tokens of the correct path. The generated hard negatives are then used during the fine-tuning stage which is trained with a path ranking object via cross entropy loss. After retraining the VLN-BERT model with the \textit{Shuffling} objective, we evaluate the model over the validation splits of the R2R dataset with no ablations to the input text or images, see results in Table \ref{table:vln_results}. We then evaluate \textit{Shuffling} model using the ablation experiments from Section~\ref{methods} over the val-unseen split to verify if this training method changed the models focus over input language during inference.

\subsection{Masked Language Modeling} 
All the models tested in this work use multi-modal transformer architectures, which are trained in part by a masked modeling objective. In this objective a randomly selected set of the input text tokens and image tokens are masked and the model must predict the original values of the masked tokens given the context of the unmasked tokens. ViLBERT~\cite{lu2019vilbert} proposed that instead of directly predicting the masked image tokens instead the model can predicting a distribution over object classes present in the masked region. Masked text tokens, however, are handled the same as in BERT where the original token is predicted directly. The masked language modeling (MLM) objective is used during stage 3 of training for VLN-BERT and randomly masks out 15\% of the tokens. 

In Section ~\ref{comp_language} found that there are a low density of spatial words such as [\textit{left, right}], see Table~\ref{table:dataset_comparison}. The lower density will result in spatial tokens being masked  significantly less than other token types during the original training procedure. As the transformer model learns the masked tokens, we hypothesize this may be limiting the agent from learning a advanced representation of spatial tokens. Therefore, to encourage the model to learn the connection between spatial words and the path, we propose to train the model for an additional 10 epochs at the end of stage 3 of the VLN-BERT pre-training schema which does the MLM. In these last 10 epochs, 100\% of spatial and directional tokens are masked over each instruction and no other tokens are masked. Additionally, no image tokens are masked creating a purely masked language modeling objective. The model is then fine-tuned with the original cross-entropy loss. We call this training method: \textit{spatial language loss (SLL)} and we evaluate the model over the validation splits of the R2R dataset with no ablation to the input tokens (no tokens are masked during inference), see Table \ref{table:vln_results}. The intention of SLL is to alleviate differences in token density and the larger weighting on nouns we can see this akin to a weighted loss or penalty corresponding to the density of the token type. Note the masking of spatial tokens only takes place during the final 10 epochs of training as VLN-BERT learns by trying to predict masked tokens and no tokens are masked at inference (Table \ref{table:vln_results}). 

\subsection{Training via Passive Data}
\label{passive-approach}
Path-ranking transformer based models for VLN rely heavily on noun tokens while disregarding other types of tokens. Directional and spatial tokens provide significant information that these models are currently not taking advantage of. To combat model reliance on tokens that describe object and rooms, we inject new data during the training stage on R2R which contains sparse references to object and room names. We programmatically generate new paths paired with new object-sparse and direction-dense instructions. This allows us to forgo the need for additional human annotations or need to train an additional speaker model~\cite{fried2018speaker}.   

\xhdr{Generation of New Instruction-Pairs.}\\
\noindent We generate the additional navigational paths using a similar strategy to \cite{anderson2018evaluation}. We first sample start-goal location pairs in the MP3D~\cite{chang2017matterport3d} training environments and then determine the shortest path between the pair via scene navigation graph. Generated paths are discarded if they are contained in the R2R dataset or if they are too short or too long. We select paths with  >5 edges and <10 edges. 

After generating new paths in the MP3D training environments we generate an instruction for each paths. We seek to generate instructions with sparse object references and create instructions in which an agent would be forced to follow spatial cues. Therefore we can programmatically generate a natural language navigation instruction based on the 3D coordinate information of the path alone. Our rule based instruction generation is composed as follows: If a path contains a turn we will add an instruction to turn to the sentence. If the agent must turn over 120\degree degrees we add the phrase "turn around" to the instruction. If the agent must turn over 30\degree degrees we add the phrase \textit{``turn (left | right)''} to the instruction. If the agent does not need to turn, a phrase from the set [\textit{``go straight'', ``go forward'', ``continue straight''}] combined with the phrase \textit{``for $x$ meters.''} will be added to the instruction; where $x$ is the number of meters before the agent must turn or stop navigating. We can determine if the path traverses stairs based on the $z$ coordinate change of the path. If the path is determined to be traversing up or down stairs the corresponding phrase \textit{``go (up|down) the stairs''} will be added to the instruction. The final phrase of each generated instruction is chosen from the set [\textit{``stop'', ``wait here''}]. We generate an additional ~6k instruction-path pairs over the training environments of MP3D. An example generated instruction in our dataset is \myquote{Go forward and walk 3 meters. Turn right, and walk one meter. Stop.} We chose to randomly select phrases out of comparable sets of phrases as a method of adding linguistic diversity.

The generated instruction-path pairs are added to the training split of the R2R dataset. We choose to add these noun-sparse instructions to the original data instead of replacing the original data because removing all nouns from all pre-training steps would likely significantly decrease performance as all objects would be novel to the agent during inference.

We retrain a VLN-BERT model with the additional data for Stage 3 using the the masked multi-modal modeling objectives. We perform fine-tuning using the original RxR train split. We evaluate this model over the validation split of R2R for the standard VLN task (Table~\ref{table:vln_results}) as well as run this model over the masking ablation experiments (Table~\ref{table:masking_newmodels}). 

\subsection{Results}

\begin{table}[!h]
\begin{center}
\caption{Results of ablation-masking experiments: Success Rate (SR $\uparrow$) on the val-unseen split of the R2R dataset. Results across different training methods for VLN-BERT~\cite{majumdar2020improving}. The first row is the models performance with no augmentation to the input language.}
\label{table:masking_newmodels}
\resizebox{\columnwidth}{!}{
\begin{tabular}{l cccc}
\toprule   
Ablation & VLN-B~\cite{majumdar2020improving}& Shuffling & Shuffling + SLL & Generated Instr
\\ \toprule  
Original Input & 55.90 & 59.00 & 60.62 & 54.53\\ 
No Language & 30.23 & 28.99 & 31.89 & 25.71 \\
No Vision   & 26.52 & 18.86 & 21.41 & 13.96 \\
\midrule
Nouns       & 43.68 & 45.76 & 43.34 & 40.27\\
Adjectives  & 55.77 & 58.66 & 60.41 & 53.64\\
Verbs       & 55.77 & 58.96 & 43.34 & 53.64\\
Objects     & 50.40 & 52.96 & 52.36 & 46.87\\
Numerical   & 55.94 & 58.83 & 60.54 & 55.98\\
Spatial     & 55.64 & 57.26 & 59.00 & 54.07\\
Left-Right  & 55.56 & 58.24 & 60.41 & 54.45\\
Swap        & 56.07 & 56.41 & 57.43 & 53.98\\
\bottomrule
\end{tabular}}
\end{center}
\end{table}

\begin{table*}[!ht]
\begin{center}
\caption{Results of the VLN task over the R2R dataset validation splits. Compares the VLN-BERT model architecture trained with different data augmentation and training objectives. Note that all models in this table were re-trained starting from Stage 2~\cite{majumdar2020improving} using ViLBERT model weights for stage 1 and 2 in order to allow fair comparison.}
\resizebox{.7\textwidth}{!}{
\label{table:vln_results}
\begin{tabular}{l ccc c ccc}
\toprule
& \multicolumn{3}{c}{val-seen}&& \multicolumn{3}{c}{val-unseen} \\
\cmidrule(l{2pt}r{2pt}){2-4} \cmidrule(l{2pt}r{2pt}){6-8}
Method  & \multicolumn{1}{c}{\scriptsize NE $\downarrow$}& \multicolumn{1}{c}{\scriptsize SR $\uparrow$}   & \multicolumn{1}{c}{\scriptsize SPL $\uparrow$}   && \multicolumn{1}{c}{\scriptsize NE $\downarrow$}& \multicolumn{1}{c}{\scriptsize SR $\uparrow$}   & \multicolumn{1}{c}{\scriptsize SPL $\uparrow$} \\ \toprule
VLN-BERT~\cite{majumdar2020improving} & 4.1093 & 66.67 & 63.09  && 4.6699 & 55.90 & 51.58	 \\ \midrule
Shuffling  & 4.2836 & 66.08 & 62.09  && 4.5222 & 59.00 & 54.36  \\
Shuffling + SLL & 4.3506 & 64.90 & 61.05 && \textbf{4.1503} & \textbf{60.62} & \textbf{56.08 }\\
R2R + Passive Data & \textbf{3.8264} & \textbf{68.04} & \textbf{63.90}  && 4.7218 & 54.53 & 50.00   \\
\bottomrule
\end{tabular}}
\end{center}
\end{table*}

\begin{figure}[!ht]
\centering
\includegraphics[width=\columnwidth]{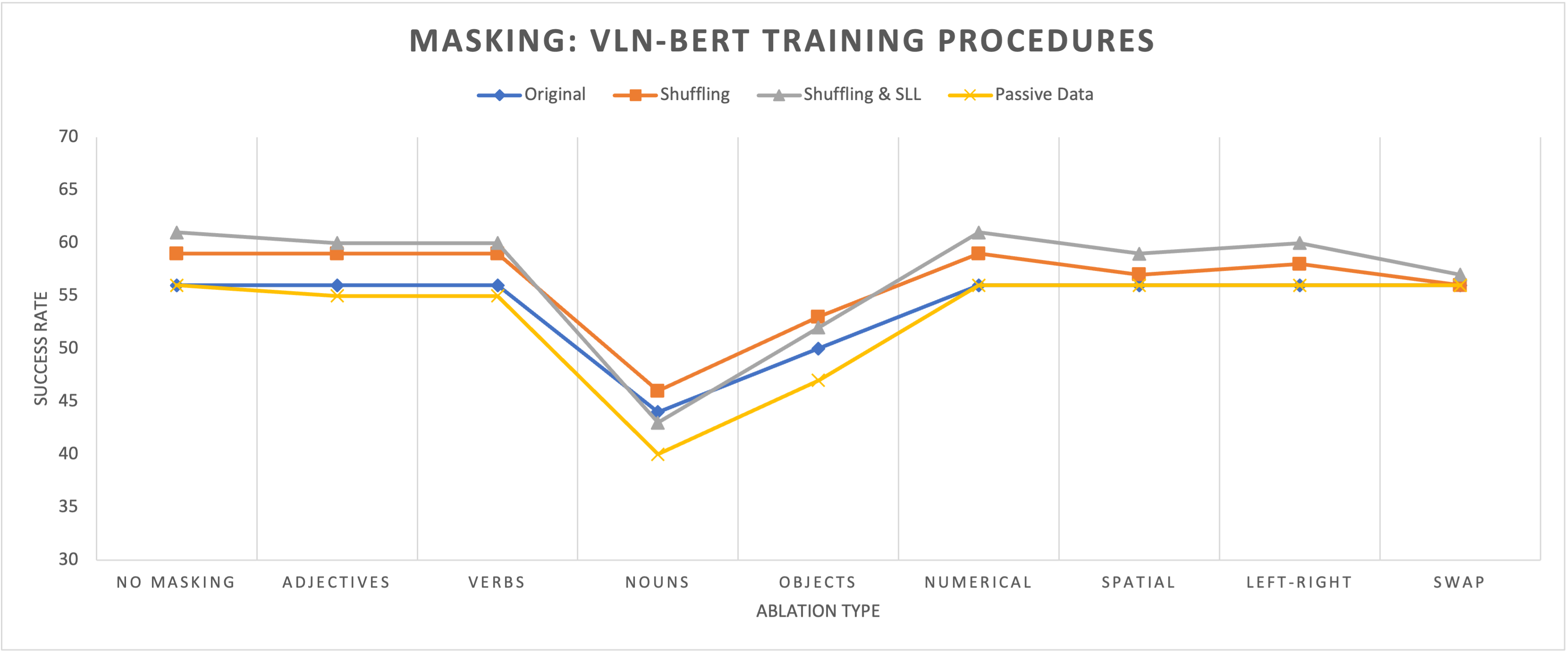}
\caption{This figure displays a visual illustration of the results of the masking experiments over the different training procedures for VLN-BERT.}
\label{masking-graph-vln-bert}
\end{figure}

\xhdr{Evaluation via Standard VLN Task.}\\
\noindent We evaluate the three modified VLN-BERT models described above via the standard VLN task using the R2R validation splits. In this experiment we do not remove or replace any of the input instruction-path pairs. We report the results of these experiments as well as the results of a standard trained VLN-BERT model in Table~\ref{table:vln_results} and the results are visualized in Figure~\ref{masking-graph-vln-bert}. For the sake of  fair comparison we train all models from the Stage 2 model checkpoint provided by ViLBert~\cite{lu2019vilbert}. In other words our training takes place in Stage 3 and 4 -- see ~\cite{majumdar2020improving} for further details. We train all models with only the changes stated above and hold all other variables constant; such as using the same random seeds, model parameters, and devices. We found that after training stage 3 and 4 of the vanilla VLN-BERT model according to training specifications outlined in~\cite{majumdar2020improving} we were unable to replicate the same numbers as reported in their paper and we observe a 3.36\% drop in SR. Via analysis and speaking with the authors of this paper we attribute the performance drop to difference in training devices and GPU driver versions. 

In Table ~\ref{table:vln_results}, we observe that the Shuffling data augmentation method increases model performance significantly up to 3.1\% absolute percentage in success rate over the original model on unseen environments. Combining the methods of Shuffling and SLL increases success rate by an additional 1.62\% on the unseen validation split. Interestingly we find that training with the additional object-sparse instructions lead to little change in model performance over the original VLN-BERT model. The Passive Data model seems to have more environment bias to the training environments as it performs the best of all methods on the split of seen environments and performs the worst of all methods on the unseen environments. We also note that training on passive training data before fine-tuning did not show significant change in performance. One direction for future work is to generate object-sparse and direction-dense instruction over a different set of environments such as the newly released a HM3D~\cite{ramakrishnan2021habitat} dataset containing 1k additional Matterport environments.\\

\xhdr{Evaluation via Ablation Framework.}\\
\noindent We evaluate the three proposed training modifications the VLN-BERT using the ablation experiments proposed in Section~\ref{methods}. Running the ablation experiments over the new models allows us to investigate if any of the modifications to training in turn changed the model's focus over text inputs. We would be very encouraged to see if the new training methods increased the model's exploitation of linguistic spatial information. We show the results from these experiments in Table~\ref{table:masking_newmodels}. 

As we can observe in Table~\ref{table:masking_newmodels}, combining the Shuffling + SLL training methods produce a model that not only achieves ~5\% absolute percentage better success rate, but it seems to demonstrate a higher reliance on spatial words than the original model. This is particularly noticeable in the results of the \textit{Swap} experiment. When `left' and `right' tokens are swapped the VLN-BERT model SR increases very slightly. In contrast, for the same experiment the Shuffling + SLL model sees a drop in SR of \textbf{3.19\%} absolute percent. Out of all path-ranking models and ablation experiments this is the \emph{largest delta} in performance outside of noun masking. This further indicates to us the Shuffling and SLL training techniques encourage the models leveraging of spatial information. 

The \textit{Passive Data} training method of adding auto-generated instructions with sparse object references and dense spatial references, interestingly, leads to negligible change in the masking experiments. We find that under all token sets except nouns, there is a <1\% change in performance. This demonstrates that the additional data is not encouraging the model to utilize none noun tokens.  

Note only some of the proposed adaptations for the training procedure are also applicable to sequential types of VLN models. Shuffling is only applicable to the path-ranking models. Spatial Language Loss and add programmatically generated data can be tested in the sequential setting. We ran both of these training adaptations with Recurrent VLN-BERT training. We found adding generated sparse-noun data seems to improve performance marginally (1.5\%). We found that training with SSL does not lead to any improvement (<.1\%). We assume the training adaptions did not increase performance significantly since the sequential models are already utilizing tokens of all parts of speech. 

\section{Conclusion}
\label{sec:conclusion}
In this work we highlight a significant difference between the capabilities of the sequential and path-ranking methods for Vision-Language Navigation (VLN). Despite being a highly studied task there remains a large gap between human performance and current SOTA performance on the VLN task. Motivated by this deficiency gap, we present a robust set of ablation experiments over the language modality that are designed with aim of investigating degree to which linguistic cues inform navigational decisions. The results of these ablation experiments expose a key limitation of VLN models trained and tested in the path-ranking paradigm. We find the most popular path-ranking VLN models heavily rely on noun and object tokens of the navigation instructions and seem to disregard or under utilize spatial instructions present in the navigation. In contrast we find that sequential VLN methods do rely on directional tokens and perform as expected under the ablations.

We posit a leading cause of the issue with path-ranking models is the insufficient diversity and size of the number of training environments in the common VLN benchmark datasets. The results suggest the use of these benchmarks for training in combination with evaluating path-ranking methods is insufficient in complexity, such that it leads to the agents taking the shortcut of learning object classification rather than learning the complex multi-modal understanding of navigation. 
MP3D has 61 training environments and R2R has 4,675 unique path-instruction pairs (training set) over these environments. In Section~\ref{passive-approach} we increase the number of unique path-instruction pairs to ~11k and observe improvement in model performance which leads us to conclude that model is failing due to size and diversity of the training environments. MP3D environments are also used for the common VLN datasets (VLN-CE, RxR, Reverie) serving as a unified benchmark. 

To further investigate this phenomena and provide solutions to alleviate this problem, we study the training procedure of a single path-ranking model, VLN-BERT. We further present findings of how modifying training objectives and data augmentation strategies effect overall VLN task performance as well as increase reliance on different linguistic cues. 
\clearpage

\bibliographystyle{ACM-Reference-Format} 
\bibliography{main}
\clearpage
\section{Appendix}
\label{sec:appendix}

\subsection{Masking on Sequential Models}

To further demonstrate that other sequential VLN models utilize directional and object tokens we add an ablation experiment over the PREVALENT model~\cite{hao2020towards}. 

\begin{table}[h]
\begin{center}
\caption{Results of ablation-masking experiments: Success Rate (SR) on the val-unseen split of the R2R dataset for two sequential VLN methods. Evaluated on standard trained SOTA VLN models. The first row is the models performance with no augmentations over instruction input.}
\label{table:masking_results}
\begin{tabular}{l cccc}
\toprule   
Ablation && PREVALENT~\cite{hao2020towards} & Rec-VLN-B~\cite{hong2021vln}
\\ \toprule    
Original Input && 58.02 & 62.75 \\ 
\midrule
Nouns       && 43.24  & 43.93 \\
Objects     && 49.38  & 54.63 \\
Spatial     && 50.52  & 49.64 \\
Left-Right  && 51.60  & 54.66 \\
Swap        && 47.87  & 46.70 \\
\bottomrule
\end{tabular}
\end{center}
\end{table}

\subsection{Counterfactuals}
\label{counterfacts}

\begin{figure}[!hb]
\centering
\includegraphics[width=.9\columnwidth]{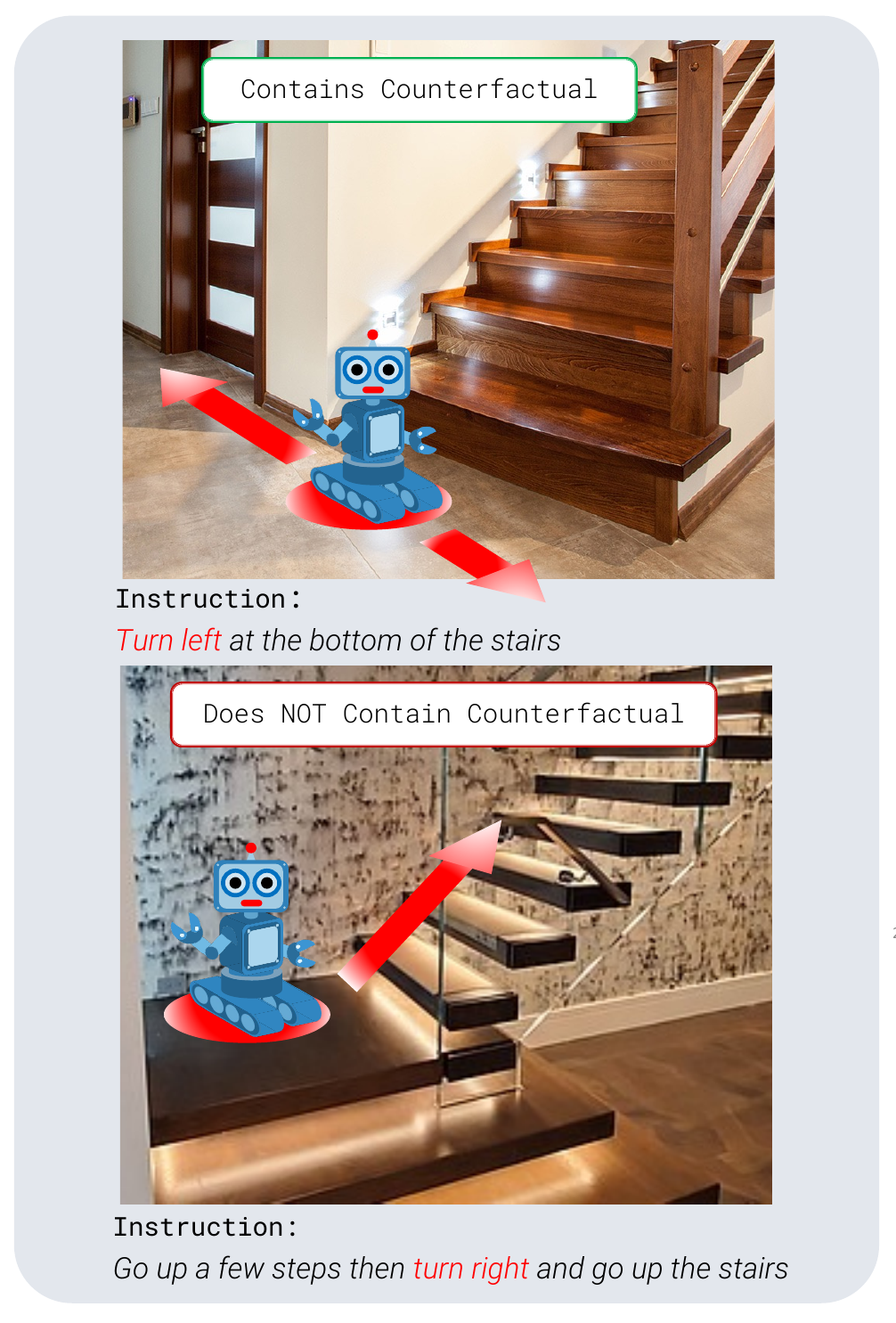}
\caption{Example of counterfactual in instruction following datasets over a specific instruction-path pair. The top image-instruction pair shows and instance where a counterfactual path exists and the phrase 'turn left' provides a essential instruction. The bottom image-instruction pair shows an example where no counter factual path exists and the phrase 'turn right' does not provide additional information.}
\label{counterfactual}
\end{figure}

As a sanity check we perform an analysis over counterfactual options in the R2R dataset. One plausible explanation for the performance not being effected in the ablations over directional tokens is that the R2R dataset contains paths that lack in counterfactual paths. In other words, when a navigation instruction states to `turn right' the environment is structured in such a way -- i.e. a hallway -- that agent has no option to turn left.  For example, imagine the navigation agent exits a room to find a hallway where only neighboring nodes are to the right of the agent -- see Figure \ref{counterfactual} for an illustration of this phenomena. We to rule out this possible explanation we perform the following sanity check and find that R2R does contain a multitude of counterfactual with 92.87\% of turns containing one or more counterfactual paths.

First for each path in val-unseen split of R2R, we annotate each time an agent turns in a path. A turn is determined to have occurred any time the heading of the agent changes by over 30\degree degrees between two adjacent nodes in the agent's path. Via this criterion, there exist an average of 1.67 turns per episode in the val-unseen split. Then at the location of each turn it is determined if there is a counterfactual path the agent could have taken instead. For instance, if the agent turned left at step $t$, could the agent taken a right turn or gone straight at step $t$ or was it bounded by the structure of the environment to take a left. We find that there are an average of 1.62 counterfactual paths per turn the agent makes. With a ratio of 92.87\% of turns containing at least one counterfactual path. These findings eliminate the possibility that directional tokens do not serve a significant role in the VLN task.

\end{document}